\documentclass{article}

\usepackage{configuration/arxiv}

\usepackage{times}
\usepackage{latexsym}

\usepackage[T1]{fontenc}

\usepackage[utf8]{inputenc}

\usepackage{microtype}

\usepackage{hyperref}
\hypersetup{
    bookmarks=true,
    bookmarksnumbered=true,
    bookmarksopen=true,
    bookmarksopenlevel=2,
    colorlinks=true,
    linkcolor=black,
    citecolor=blue,
    urlcolor=black
}

\usepackage{url}                    %
\usepackage{booktabs}               %
\usepackage{amsfonts}               %
\usepackage{nicefrac}               %
\usepackage{xcolor}                 %
\usepackage{algorithm}
\usepackage{algorithmic}
\usepackage{graphicx}
\usepackage[flushleft]{threeparttable}
\usepackage{float}
\usepackage{makecell}
\usepackage{multirow}
\usepackage{xspace}
\usepackage{natbib}
\usepackage{enumitem}
\usepackage[font=small]{caption}
\usepackage{autobreak}
\usepackage{sidecap}
\usepackage{wrapfig}
\usepackage{bbding}
\usepackage[toc, page, header]{appendix}
\usepackage{tikz}
\usepackage{colortbl}
\usepackage{xcolor}
\usepackage{pifont}
\usepackage{mdframed}
\usepackage[capitalize,noabbrev,nameinlink,sort]{cleveref}
\usepackage{tcolorbox}
\tcbuselibrary{breakable}

\usepackage{amsmath}
\usepackage{amssymb}
\usepackage{mathtools}
\usepackage{amsthm}
\usepackage{amsfonts}       

\theoremstyle{plain}
\newtheorem{theorem}{Theorem}[section]

\theoremstyle{definition}
\newtheorem{definition}[theorem]{Definition}

\theoremstyle{remark}
\newtheorem{remark}[theorem]{Remark}

\providecommand{\abs}[1]{\left\lvert#1\right\rvert}
\providecommand{\norm}[1]{\left\lVert#1\right\rVert}

\providecommand{\R}{\mathbb{R}} %

\providecommand{\vv}{\mathbf{v}}

\providecommand{\cA}{\mathcal{A}}

\providecommand{\cD}{\mathcal{D}}

\providecommand{\cP}{\mathcal{P}}

\providecommand{\cS}{\mathcal{S}}
\providecommand{\cT}{\mathcal{T}}

\providecommand{\cW}{\mathcal{W}}

\newcommand{\algopt}{\textsc{MorphAgent}\xspace}

\newcommand{\cmark}{\ding{52}} 
\newcommand{\xmark}{\ding{55}} 
\newcommand{\hmark}{\ding{52}\rotatebox[origin=c]{-9.2}{\kern-0.7em\ding{55}}} 

\title{\algopt: Empowering Agents through Self-Evolving Profiles and Decentralized Collaboration\looseness=-1}

\author{
    Siyuan Lu$^{1, 4, \ast}$, Jiaqi Shao$^{2, 3,}\thanks{Equal contributions. Work was done during Siyuan's visit to Westlake University, and Jiaqi's visit to Duke Kunshan University.}$ \,, Bing Luo$^{2, \dagger}$, Tao Lin$^{4, 5,}\thanks{Corresponding author.}$ \\
    $^1$Queen Mary University of London\\
    $^2$Duke Kunshan University\\
    $^3$Hong Kong University of Science and Technology\\
    $^4$School of Engineering, Westlake University \\
    $^5$Research Center for Industries of the Future, Westlake University
}

\begin{document}

\maketitle

\begin{abstract}
    Large Language Model (LLM) based multi-agent systems (MAS) have shown promise in tackling complex tasks, but often rely on predefined roles and centralized coordination, limiting their adaptability to evolving challenges. This paper introduces \algopt,
    a novel \textbf{Autonomous}, \textbf{Self-Organizing}, and \textbf{Self-Adaptive Multi-Agent System} for \textit{decentralized} agent collaboration that enables agents to \textit{dynamically evolve their roles and capabilities}.
    Our approach employs self-evolving agent profiles, optimized through three key metrics, guiding agents in refining their individual expertise while maintaining complementary team dynamics.
    \algopt implements a two-phase process: a \textbf{Profile Update} phase for profile optimization, followed by a \textbf{Task Execution} phase where agents continuously adapt their roles based on task feedback.
    Our experimental results show that \algopt outperforms existing frameworks in terms of task performance and adaptability to changing requirements, paving the way for more robust and versatile multi-agent collaborative systems.
\end{abstract}

\section{Introduction}\label{sec:introduction}

\begin{wrapfigure}{r}{0.45\linewidth}
    \centering
    \includegraphics[width=0.7\linewidth]{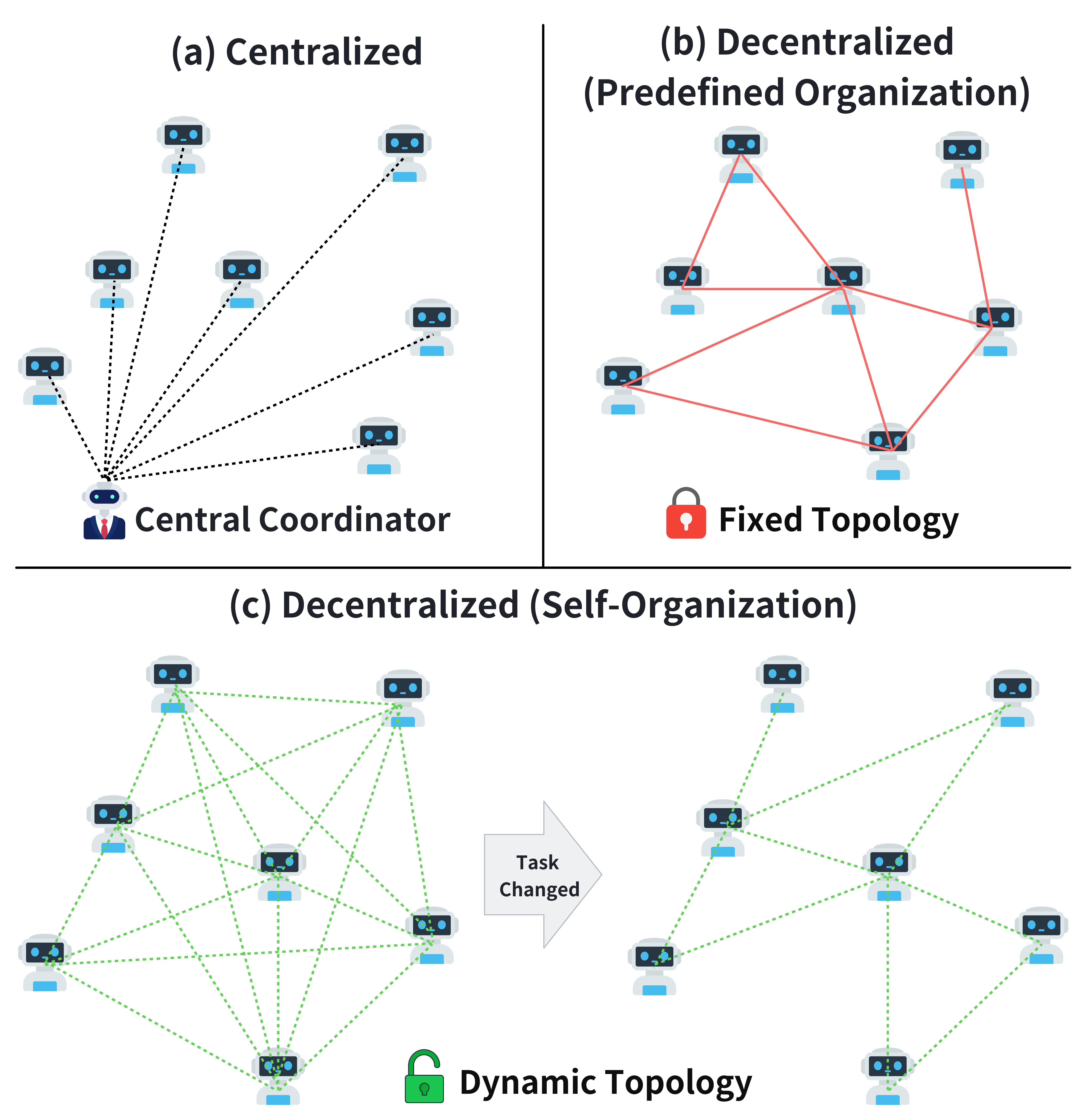}
    \vspace{-0.5em}
    \caption{\small
        \textbf{Comparison of LLM-based MAS architectures:}
        (a) Centralized systems with a single coordinator;
        (b) Decentralized systems with fixed topologies;
        (c) Decentralized systems with dynamic yet self-organizing topologies.
    }\label{fig:intro}
    \vspace{-1.25em}
\end{wrapfigure}

The rapid advancement of Large Language Models (LLMs)~\citep{achiam2023gpt, touvron2023llamaopenefficientfoundation} has ushered in a new era of artificial intelligence, enabling the creation of sophisticated AI agents capable of tackling complex tasks across diverse domains~\citep{githubGitHubYoheinakajimababyagi, githubGitHubSignificantGravitasAutoGPT}.
As these AI systems become more intricate, there has been notable progress in developing communicative agents for completing collaborative tasks together~\citep{ijcai2024p0890,wang2024survey,qian2023communicative,hong2024metagpt}.
This collaborative approach, known as Multi-Agent Systems (MAS)~\citep{han2024llm}, has shown great promise in addressing challenges that are too complex or diverse for single-agent systems~\citep{hong2024metagpt, liu2023training}.
\looseness=-1

In the meanwhile, natural systems demonstrate remarkable forms of collective behavior through decentralized interactions among individuals~\citep{beni1993swarm}.
From ant colonies coordinating complex foraging patterns to bird flocks executing synchronized aerial maneuvers,
these systems achieve sophisticated group behaviors without central control~\citep{duan2023animal}.
The success of these natural swarms lies in their ability to exhibit emergent intelligence that surpasses individual capabilities,
offering valuable insights for designing more robust and adaptable multi-agent systems~\citep{rosenberg2020artificial}.

\begin{table*}[!t]
    \centering
    \small
    \tabcolsep=0.01\linewidth
    \caption{
    \textbf{Comparison of methods on three key properties}: Individual Autonomy, Self-Organization, and Self-Adaptability.
    \cmark~indicates full support, \xmark~indicates no support, and \hmark~indicates partial support.
    Some approaches~[\citenum{chen2024agentverse,guo2024embodied,hu2024self}], as illustrated in~\cref{fig:intro} (a), attempt \textit{Self-Organization} by using specialized agents for task coordination. However, this centralized orchestration restricts the flexibility and robustness of agent-driven collaboration.
    Other methods~[\citenum{zhuge2024gptswarm,zhang2025aflow}], shown in~\cref{fig:intro} (b), use predefined mechanisms to structure collaboration workflows for different tasks, limiting emergent organization.
    \looseness=-1
    }\label{tab:comparison}
    \begin{tabular}{lccc}
        \toprule
        \textbf{Method}                           & \textbf{\makecell{Individual Autonomy}} & \textbf{\makecell{Self-Organization}} & \textbf{\makecell{Self-Adaptability}} \\ \midrule
        MetaGPT~\citep{hong2024metagpt}           & \xmark                                  & \xmark                                & \xmark                                \\
        GPTSwarm~\citep{zhuge2024gptswarm}        & \xmark                                  & \hmark                                & \xmark                                \\
        AFLOW~\citep{zhang2025aflow}              & \xmark                                  & \hmark                                & \xmark                                \\
        AgentVerse~\citep{chen2024agentverse}     & \cmark                                  & \hmark                                & \xmark                                \\
        Criticize-Reflect~\citep{guo2024embodied} & \cmark                                  & \hmark                                & \hmark                                \\
        EvoMAC~\citep{hu2024self}                 & \cmark                                  & \hmark                                & \hmark                                \\  \midrule

        \algopt                                   & \cmark                                  & \cmark                                & \cmark                                \\ \bottomrule
    \end{tabular}
    \vspace{-1.em}
\end{table*}

Inspired by the remarkable emergent intelligence observed in these swarm systems~\citep{brambilla2013swarm,rosenberg2020artificial,nguyen2024swarm}, we distill three fundamental principles that serve as the foundation for designing decentralized and adaptive multi-agent systems:
\begin{enumerate}[nosep, leftmargin=12pt]
    \item \textbf{Individual Autonomy:} each agent processes local information and independently determines its actions during collaboration, rather than following rigid workflows.
          \looseness=-1
    \item \textbf{Self-Organization:} agents spontaneously form collaborate patterns through simple interaction rules, including information sharing, behavioral alignment, and mutual adaptation.
          This dynamic collaboration structures is crucial for handling unpredictable environments.
          \looseness=-1
    \item \textbf{Self-Adaptability:} agents dynamically adjust their roles, responsibilities, and behaviors to ensure resilience in dynamic environments.
\end{enumerate}

However, existing Multi-Agent Systems (MAS) fall short of fully embodying the three key principles outlined in~\cref{tab:comparison}. First, \textit{individual autonomy is compromised} in systems such as GPTSwarm~[\citenum{zhuge2024gptswarm}], AFLOW~[\citenum{zhang2025aflow}], and MetaGPT~[\citenum{hong2024metagpt}], where rigid workflows predetermine agent roles and actions, limiting the agents' independent decision-making capabilities.
Although some methods incorporate aspects of self-organization, they often rely on centralized task coordinators—as seen in AgentVerse~[\citenum{chen2024agentverse}], Criticize-Reflect~[\citenum{guo2024embodied}], and EvoMAC~[\citenum{hu2024self}]—or predefined collaboration structures~[\citenum{zhuge2024gptswarm,zhang2025aflow}].
These constraints \textit{hinder the emergence of truly decentralized and adaptive teamwork}. Finally, \textit{self-adaptability remains limited} across most MAS. In many cases, agent roles are either fixed or can only be modified through external coordination rather than through intrinsic, self-driven evolution~[\citenum{chen2024agentverse, guo2024embodied, hu2024self}].

To address these limitations, as illustrated in~\cref{fig:intro} (c), we propose \algopt, a robust, scalable, and adaptive decentralized system capable of addressing complex and dynamic challenges.
Our framework enables each LLM-based agent to independently adapt to dynamic tasks and environments, exemplifying \emph{individual Autonomy}.
Additionally, \algopt introduces a dynamic profile mechanism, which defines and continuously evolves each agent's roles, capabilities, and expertise based on interactions and task demands without a centralized control.
To ensure \emph{self-organization} and \emph{self-adaptability}, we introduce three metrics---namely role clarity, team diversity, and capability matching---to enhance collaboration and facilitate continuous adaptation to changing environments.
Through dynamic profile optimization and agent collaboration, \algopt achieves swarm intelligence in decentralized multi-agent systems.

\noindent\textbf{Contributions}. We summarize the key contributions:
\begin{itemize}[leftmargin=12pt, nosep]
    \item We propose \algopt, a decentralized multi-agent framework with the properties of individual autonomy, self-organization, and self-adaptability.
          \looseness=-1
    \item We present an innovative dynamic profile mechanism incorporating three metrics, which facilitates the continuous evolution of agents' roles and capabilities, thereby enabling effective self-organization and collaboration in dynamic environments.\looseness=-1
    \item Comprehensive empirical results demonstrate that our method consistently outperforms SOTA methods across various tasks.
          Notably, in a challenging environment with high failure rates (0.8), which means that the agent has a 80\% probability of failing to act,
          our method maintains robust performance (40.10 $\sim$ 54.00\% across tasks) while baselines degrade significantly (1.43 $\sim$ 19.52\%).
\end{itemize}
\section{Related Work}
\label{sec:related_work}
\paragraph{LLM-based Multi-Agent Systems (MAS).}
The emergence of Large Language Model~\citep{achiam2023gpt, touvron2023llama} has led to LLM-based autonomous agent capable of tackling complex tasks, like BabyAGI~\citep{githubGitHubYoheinakajimababyagi} and AutoGPT~\citep{githubGitHubSignificantGravitasAutoGPT}.
However, single LLM agent often struggle with cooperative work, such as software engineering~\citep{jimenez2024swebench}.
To address these limitations, recent studies have proposed LLM-based MAS~\citep{han2024llm,zhou2023sotopia}, where multiple AI agents collaborate on solving complex tasks. \looseness=-1

However, current approaches often rely on predefined roles~\citep{li2023camel,hong2024metagpt}, centralized coordination~\citep{chen2024agentverse}, or rigid organizational structures~\citep{hong2024metagpt, qian2024chatdev}.
For example, CAMEL~\citep{li2023camel} and ChatEval~\citep{chan2023chateval} employ agents with predefined roles through role-playing, but struggle to adapt to tasks requiring unforeseen skills.
MetaGPT~\citep{hong2024metagpt} implements human workflow in rigid organizational structures, showing improvements in code-generation but lacking generalization to other tasks like writing a research paper.
Our work addresses these limitations by initializing agents homogeneously without predefined roles or collaboration structures, allowing them to naturally develop cooperation through interaction.

\paragraph{Organization optimization for MAS.}
Recent research in LLM-based MAS has focused on optimizing organizational structures~\citep{guo2024embodied, zhuge2024gptswarm} and enhancing agent performance~\citep{zhang2024proagent} to reduce communication costs and increase team efficiency.
Approaches like AgentVerse~\citep{chen2024agentverse}, Criticize-Reflect~\citep{guo2024embodied}, and EvoMAC~\citep{hu2024self} rely on centralized mechanisms, where a single role or a subset of agents monitor and evaluate the system's overall performance.
While effective in certain scenarios, these centralized methods may face scalability issues and potential bottlenecks in large-scale MAS.
Our decentralized approach leverages LLM-based agents' self-reflection capabilities~\citep{madaan2023selfrefine, shinn2023reflexion, renze2024self}, and enables agents to dynamically adjust their responsibilities based on the change of tasks for better adaptivity.
\looseness=-1

\paragraph{Standard Operating Procedure (SOP) based MAS.}
Another line of research has explored more structured collaboration methodologies in LLM-based multi-agent systems.
SOP-based approaches such as AgentCoder~\citep{huang2023agentcoder} and MetaGPT~\citep{hong2024metagpt} have demonstrated performance gains through standardized workflows.
GPTSwarm~\citep{zhuge2024gptswarm} and AFLOW~\citep{zhang2025aflow} extends these methods by modeling agents as subnets of action nodes and constructing workflows for diverse tasks.
Although SOP-based approaches offer efficient coordination for adhering to predefined procedures in specific scenarios, they lack flexibility.
Our framework empowers multi-agent collaboration with autonomous planning capabilities of advanced LLM-based agents~\citep{huang2022language, guan2023leveraging, wang2023describe}: instead of relying on rigid SOPs, it facilitates the dynamic development of collaborative strategies and efficient role adaptation, improving overall performance and robustness.
\begin{figure*}[!t]
    \centering
    \includegraphics[width=0.9\textwidth]{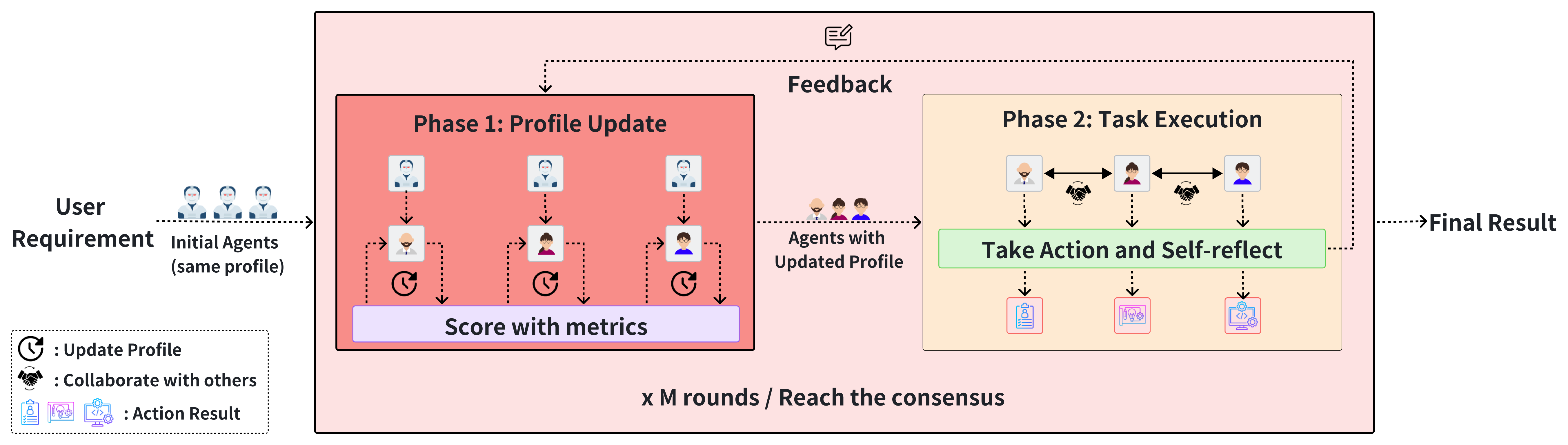}
    \caption{
        \textbf{Pipeline of \algopt.}
        Agents start from user requirements, undergo a profile update phase to optimize profiles based on metrics (terminating after a set number of rounds or upon reaching the metric threshold), and then proceed to the task execution phase, where profiles are updated iteratively.
        The task execution phase ends when consensus is reached or required rounds are completed, with feedback loops ensuring continuous adaptation.
        \looseness=-1
    }
    \vspace{-1.5em}
    \label{fig:framework}
\end{figure*}

\section{\algopt}\label{sec:method}

Our framework, \algopt, enables decentralized and adaptive multi-agent collaboration by integrating three key components.
First, we define the fundamental agent properties (c.f.~\cref{sec:properties}), introducing dynamic profiles that allow agents to autonomously adjust their roles and behaviors.
Next, we present the Decentralized Collaboration Framework (in \cref{sec:autonomous_collab}), which orchestrates agent interactions through iterative and decentralized coordination.
Finally, we introduce profile evaluation and optimization metrics (c.f.~\cref{sec:metrics}), which ensure role clarity, team diversity, and effective task alignment, facilitating robust collaboration.

\subsection{System Properties} \label{sec:properties}
The LLM-based agent serves as the fundamental building block of our framework,
capable of autonomous reasoning, decision-making, and interaction with other agents.
To facilitate systematic analysis and optimization of agent behavior in a MAS, we begin by a concrete definition of agent's profile in a MAS.
\begin{definition}[Agent's Profile]
    Each agent $a_i \in \mathcal{A}$ in a MAS is associated with a \textit{profile} $p_i$, which encodes its roles, capabilities, and interaction preferences.
    Formally, the profile is represented as $p_i = (f_i, c_i, r_i)$, where $f_i$ denotes the agent's functional capabilities (e.g., specific skills), $c_i$ captures contextual or task-specific adjustments, and $r_i$ defines the rules for interaction with other agents.
\end{definition}
\begin{remark}
    Profiles can be either \emph{immutable}, remaining constant throughout the system's operation, or \emph{dynamic}, evolving over time based on local observations, task requirements, and peer interactions.
    Dynamic profiles enable agents to adapt to changing environments, unlike the static, immutable profiled used in traditional MAS~\citep{li2023camel,hong2024metagpt}.
\end{remark}

Specifically, we focus on the dynamic profile, which evolves over time based on local observations, task requirements, and peer interactions.
The dynamic profile mechanism is defined as follows:

\begin{definition}[Dynamic Profile Mechanism]
    \label{def:dynamic_profile}
    The \textit{Dynamic Profile Mechanism} enables agents to autonomously refine their roles and collaboration strategies by updating their profiles based on evolving observations, past interactions, and task demands.
    If agents operate with \emph{Individual Autonomy}, they can independently determine adjustments to their profiles.
    Formally, at time $t$, an agent $a_i \in \mathcal{A}$ updates its profile $p_i^t$ as:
    \begin{align*}
        p_i^t = \psi(o_i^t, p_i^{t-1}, \{\alpha_j^t\}_{j \in \mathcal{N}_i^t}) \,,
    \end{align*}
    where $o_i^t$ represents local observations, $p_i^{t-1}$ is the previous profile, and $\{\alpha_j^t\}_{j \in \mathcal{N}_i^t}$ are actions from neighboring agents.
    The function $\psi$, implemented using an LLM, enables agents to dynamically adjust their roles.
\end{definition}

Building upon the dynamic profile, we define three core principles derived from natural systems, namely, \emph{Individual Autonomy}, \emph{Self-Organization}, and \emph{Self-Adaptability}, as defined below:
\begin{definition}[Property \# 1: Individual Autonomy]
    \label{def:individual_autonomy}
    Each agent $a_i \in \mathcal{A}$ at time $t$ operates independently, without relying on a central controller or predefined workflows.
    More precisely, the action of $\alpha_i^t$ of agent $a_i$ is sampled from a policy function $\pi$, based on the local observations $o_i^t$, historical states $h_i^{t-1}$, and its dynamic profile $p_i^t$:
    \begin{align*}
        \alpha_i^t \sim \pi(o_i^t, h_i^{t-1}, p_i^t) \,,
    \end{align*}
    where
    policy $\pi$ can be implemented as a LLM, by mapping these inputs to a probability distribution over actions.
\end{definition}
\begin{remark}
    Unlike existing MAS constrained to a limited set of predefined actions or skill sets~\citep{li2023camel, hong2024metagpt}, $\alpha_i^t$ can involve high-level reasoning or complex behaviors.
    By generating diverse actions through $\pi$, each agent exemplifies autonomy, enabling flexible responses to evolving tasks and environments.
    \looseness=-1
\end{remark}

\begin{definition}[Property \# 2: Self-Organization]
    \label{def:self_organization}
    A MAS exhibits \textit{Self-Organization} if agents independently establish and adapt their interaction patterns without relying on predefined structures or centralized coordination.
    Such interaction topology evolves dynamically based on the changes in agent states, task requirements, and environmental conditions, as shown in~\cref{fig:intro}c.
\end{definition}
\begin{remark}
    Each agent independently determines which other agents to collaborate with based on its own observations, dynamic profile, and past interactions. These decisions follow adaptive interaction rules that guide how connections are formed and adjusted over time.
    This decentralized mechanism enables the system to dynamically reconfigure its topology in real time, adapting to changes in tasks or resources.
    Unlike centralized systems relying on a single coordinator (\cref{fig:intro}a) or decentralized systems with fixed topologies (\cref{fig:intro}b), self-organizing systems preserve scalability and adaptability in dynamic environments.
\end{remark}

\begin{definition}[Property \# 3: Self-Adaptability]
    \label{def:self_adaptability}
    A MAS exhibits \textit{Self-Adaptability} if agents can dynamically adjust their roles and interactions in response to evolving observations, past states, and peer behaviors. This capability ensures that agents continuously refine their expertise and collaboration strategies, allowing the system to remain flexible and resilient in dynamic environments.
\end{definition}

\paragraph{Summary.}
\algopt utilizes the dynamic profile mechanism to integrate \emph{Individual Autonomy} into profile evolution, embodying \emph{Self-Adaptability} by enabling agents to continuously refine their roles and interactions.
This adaptation process further promotes \emph{Self-Organization}, as agents collectively adjust collaboration structures without a centralized control (see~\cref{sec:case_study} for details).
\looseness=-1

\subsection{Decentralized Collaboration Framework}
\label{sec:autonomous_collab}

In this section, we introduce our \textit{Decentralized Collaboration Framework}, which orchestrates agent interactions through an iterative process.
The evaluation and optimization of profiles are supported by~\cref{sec:metrics}, while~\cref{fig:framework} illustrates how these two key phases are executed.

The framework employs autonomous agents equipped with \emph{dynamic profile mechanism} (\cref{def:dynamic_profile}) to enable seamless coordination.
Each agent in the framework employ LLMs (e.g., GPT-4~\citep{achiam2023gpt}) to execute an iterative \emph{Observe-Think-Act} cycle:
(1) observing their environment and peers' actions,
(2) planning via frameworks like ReAct~\citep{yao2022react} and Reflexion~\citep{shinn2023reflexion},
and (3) acting based on \textit{dynamic profiles}.
The agent profile is continuously updated through task feedback and peer interactions, enabling skill recalibration and fostering Individual Autonomy.
\looseness=-1

The framework operates in two key phases, ensuring robust collaboration in dynamic environments (c.f.~\cref{fig:framework}):
\looseness=-1

\begin{wraptable}{r}{0.45\linewidth}
    \centering
    \scriptsize
    \tabcolsep=0.01\linewidth
    \caption{\small Examples of feedback prompts for different profile evaluation cases.}
    \vspace{-1em}
    \label{tab:profile_prompts}
    \begin{tabular}{p{2.3cm}p{5.0cm}}
        \toprule
        \textbf{Cases}            & \textbf{Example Feedback Prompt}                                                                                              \\
        \midrule
        \textbf{Initial Profile}  & \emph{``Refine your role description to focus on key skills and eliminate ambiguity.''}                                       \\
        \midrule
        \textbf{Improved Profile} & \emph{``Your profile clarity has improved. Keep refining it.''}                                                               \\
        \midrule
        \textbf{Degraded Profile} & \emph{``Your profile's performance has degraded. Revisit recent updates to improve clarity and better align with the task.''} \\
        \midrule
        \textbf{Similar Profiles} & \emph{``Your profile is too similar to others. Highlight unique skills or responsibilities.''}                                \\
        \bottomrule
    \end{tabular}
    \vspace{-1.5em}
\end{wraptable}

\textbf{Phase 1: Profile Update.}
Agents begin with identical or minimally specified profiles, which are iteratively refined using three key metrics (see~\cref{sec:metrics} for details).
Each iteration, agents receive adaptive feedback prompts based on \textit{the changes of their metric scores} compared to the previous iteration.
The feedback prompts are designed for four different cases for score changes:
(i) \emph{initial profile}, where profiles typically lack clear descriptions and task alignment;
(ii) \emph{improved profile}, when scores \textit{increase}, reinforcing positive refinements;
(iii) \emph{degraded profile}, where scores \textit{decrease}, prompting rollback suggestions; and
(iv) \emph{similar profiles}, which indicate a lack of differentiation among agents, leading to recommendations for better role specialization.

\cref{tab:profile_prompts} provides concrete examples of how prompts are generated for each case, illustrating the refinement process.
This process reduces ambiguities, fosters diversity, and aligns roles with task demands.
Profile updates continue until stabilization or meeting stopping criteria (see examples in~\cref{tab:profile_prompts} and full details in~\cref{app:profile-optimization,app:profile-prompts}).
After the profile update phase, the agents are equipped with optimized profiles, and are ready for the task execution phase.

\textbf{Phase 2: Task Execution.}
With optimized profiles, agents execute task iteratively:
each agent \textit{observes} the environment and task state, \textit{decides} whether to execute or skip (the “skip” action is designed to terminate the agent’s task execution. When all agents choose “skip,” they reach a consensus on the final answer.) its assigned subtask, and --- if it acts --- \textit{logs} the resulting outcomes.
Agents re-enter Phase 1 upon receiving new feedback (e.g., partial failures) or detecting changes in task requirements or environmental conditions.
This dual-level adaptation ensures (1) \emph{individual} role refinement through profile updates, and (2) \emph{collective} adjustments to collaboration patterns.
\looseness=-1

In our implementation, the initial Profile Update phase runs for up to five rounds, with an early stopping mechanism:
if all three metrics (RCS, RDS, TRAS) improve by more than $0.1$, the process ends early.
After the initial phase, subsequent updates require only one round as a lightweight adjustment in response to task changes

By alternating between profile refinement (Phase 1) and task execution (Phase 2), the framework integrates \emph{self-adaptivity} and \emph{self-organization} between agents, ensuring coherence with evolving task objectives.
\looseness=-1

\begin{table*}[!t]
    \centering
    \tabcolsep=0.01\linewidth
    \caption{\small
        \textbf{Metrics for Evaluating and Optimizing Agent Profiles.}
        This table presents the three key metrics used for profile optimization: Role Clarity Score (RCS), Role Differentiation Score (RDS), and Task-Role Alignment Score (TRAS).
        Each metric is described with its core intuition, key components, and a detailed explanation.
    }
    \label{tab:metrics_compact}
    \resizebox{\linewidth}{!}{%
        \begin{tabular}{@{}p{1cm}p{5cm}p{4.8cm}p{6.5cm}@{}}
            \toprule
            \textbf{Metric}                                                                                                                                                & \textbf{Intuition}                                                                                                                & \textbf{Key Components}                       & \textbf{Explanation}                                                                                                                  \\
            \midrule
            \multirow{3}{*}{RCS}                                                                                                                                           &
            \multirow{3}{*}{\parbox[t]{4.5cm}{Defines agent unambiguous profiles by combining clear grammatical structure, non-repetitive language, and relevant skills.}} &
            Syntactic Complexity (DEP)                                                                                                                                     & Measures structural richness using dependency parsing. Profiles with deeper syntactic trees score higher than flat statements.                                                                                                                                                                                            \\
            \cmidrule(lr){3-4}
                                                                                                                                                                           &                                                                                                                                   & Lexical Diversity (ENT)                       & Computes the entropy of word usage to penalize repetition for semantically diverse language.                                          \\
            \cmidrule(lr){3-4}
                                                                                                                                                                           &                                                                                                                                   & Skill Relevance (SKILL)                       & Evaluates alignment with task-specific expertise by comparing profile tokens to skill prototypes and identifying skill-bearing nouns. \\
            \midrule
            \multirow{1}{*}{RDS}                                                                                                                                           &
            \multirow{1}{*}{\parbox[t]{4.5cm}{Promotes specialization by                                                                                                                                                                                                                                                                                                                                                                                                                               \\encouraging diversity.}}                                                                  &
            Pairwise-profile Dissimilarity                                                                                                                                 & Quantifies how distinct an agent's profile is from others.                                                                                                                                                                                                                                                                \\
            \midrule
            \multirow{2}{*}{TRAS}                                                                                                                                          &
            \multirow{2}{*}{\parbox[t]{4.5cm}{Matches agents' roles and capabilities with task requirements.}}                                                             &
            Profile-task Similarity (S$_{\mathrm{sim}}$)                                                                                                                   & Measures how closely task requirements linguistically align with agent profiles using embeddings from pretrained language models.                                                                                                                                                                                         \\
            \cmidrule(lr){3-4}
                                                                                                                                                                           &                                                                                                                                   & Capability Compatibility (S$_{\mathrm{cap}}$) & Ensures agents' capability match task's complexity.\looseness-1                                                                       \\
            \bottomrule
        \end{tabular}%
    }
    \vspace{-1em}
\end{table*}

\subsection{Profile Optimization: Challenges and Solutions}
\label{sec:metrics}
In dynamic and complex environments, resolving \textit{role ambiguity} is crucial to clarify agent responsibilities and enhance their profiles.
With well-described agent profiles, fostering \textit{diversity} equips the system to address multifaceted problems, while \textit{alignment} between agent capabilities and task requirements ensures effectiveness in executing complex tasks.

To enhance the \emph{self-adaptability} of MAS, we introduce three key metrics for evaluating and optimizing each agent's profile: Role Clarity Score (RCS), Role Differentiation Score (RDS), and Task-Role Alignment Score (TRAS). These metrics serve distinct but complementary purposes: RCS measures how clearly an agent’s role is defined, RDS captures the diversity of roles across the team, and TRAS assesses how well an agent’s role aligns with task requirements. Together, they guide dynamic profile optimization to ensure effective collaboration. Intuitions and key components of each metric are summarized in~\cref{tab:metrics_compact}.
\looseness=-1

\begin{definition}[Role Clarity Score (Informal)]
    For an agent $a \in \cA$ with profile $p \in \cS$, where $\cS$ is the set of all possible profile strings, role clarity considers the syntactic structure, lexical diversity, and skill relevance of the profile, which can be defined as:
    \[
        \text{\small $\mathrm{RCS}(a) = \beta_1 \cdot \operatorname{DEP}(p) + \beta_2 \cdot \operatorname{ENT}(p) + \beta_3 \cdot \operatorname{SKILL}(p) \,$}
    \]
    where $\beta_1 + \beta_2 +\beta_3 = 1$, and
    \begin{itemize}[leftmargin=12pt, nosep]
        \item $\operatorname{DEP}: \cS \to [0,1]$ is the dependency score, measuring syntactic complexity.
        \item $\operatorname{ENT}: \cS \to [0,1]$ is the entropy score, quantifying lexical diversity.
        \item $\operatorname{SKILL}: \cS \to [0,1]$ is the skill score, measuring relevance to skill-related concepts.
              \looseness=-1
    \end{itemize}
\end{definition}
We defer the detailed implementation of RCS to \cref{app:metrics_implementation}, e.g., the formal definition of each component.

\begin{definition}[Role Differentiation Score]
    Let $\cA = \{a_1, \ldots, a_n\}$ be a set of $n$ agents, with profiles $\cP = \{p_1, \ldots, p_n\}$.
    The role differentiation of $\cA$ measures the average dissimilarity between agent profiles, which can be defined as: \looseness=-1
    \[
        \mathrm{RDS} = h\left(\frac{2}{n(n-1)} \sum_{1 \leq i < j \leq n} d(a_i, a_j)\right) \,,
    \]
    where $d(a_i, a_j) = 1 - \frac{e(p_i) \cdot e(p_j)}{ \norm{ e(p_i) } \norm{ e(p_j) } } $ is the dissimilarity between agents $a_i$ and $a_j$ measured by
    the embeddings of
    their profiles $p_i$ and $p_j$, and $h$ is a sigmoid function to normalize the score.
\end{definition}

\begin{definition}[Task-Role Alignment Score (Informal)]
    Given a task $T \in \cT$ and a set of agent profiles $\cP = \{p_1, \ldots, p_n\}$, task-role alignment investigates how well the agents' roles align with the task requirements, which is defined as:
    \[
        \mathrm{TRAS} = \alpha \cdot S_{\mathrm{sim}}(T, \cP) + (1-\alpha) \cdot S_{\mathrm{cap}}(T, \cP) \,,
    \]
    where:
    \begin{itemize}[leftmargin=12pt, nosep]
        \item \textit{Semantic Similarity ($S_{\mathrm{sim}}$)}: Measures alignment between task requirements and agent expertise by computing the average similarity score between task descriptions and agent profiles.

        \item \textit{Capability Compatibility ($S_{\mathrm{cap}}$)}: Evaluates how well the agent capabilities match the task's complexity demands.
    \end{itemize}
\end{definition}
We defer the implementation details of TRAS to \cref{app:metrics_implementation}, e.g., the formal construction for $S_{\mathrm{sim}}$ and $S_{\mathrm{cap}}$.
\looseness=-1

These three scores jointly offer a comprehensive basis for assessing and refining agent roles, thus enabling flexible and adaptive teamwork.
\cref{app:metrics_implementation} details the practical implementation of each metric, and \cref{app:profile-analysis} demonstrates their effectiveness through quantitative evaluations on various agent configurations and task scenarios.
\begin{figure*}[!t]
    \centering
    \includegraphics[width=0.9\textwidth]{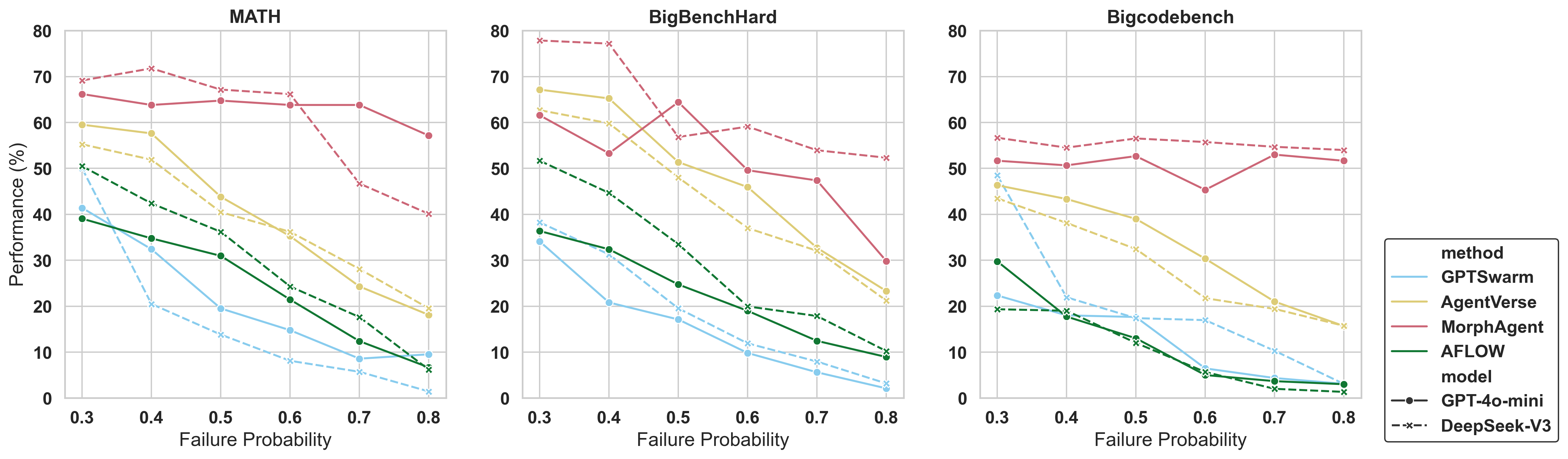}
    \caption{\small
        \textbf{Comparison with state-of-art baseline methods}:
        Our approach consistently outperforms baseline methods across all three benchmark tasks (Code Generation, General Reasoning, and Mathematical Reasoning) in multiple LLM backbones.
    }
    \label{fig:baseline_comparison}
\end{figure*}

\section{Experiments}\label{sec:exp}

In this section, we comprehensively evaluate our multi-agent collaboration framework \algopt from three perspectives.
First, in~\cref{sec:compare_baselines}, we compare \algopt with existing MAS approaches on benchmark tasks, assessing its robustness under varying conditions.
Next, in~\cref{sec:ablation_study}, we conduct an ablation study to examine the impact of profile optimization metrics and scalability with increasing agent numbers.
Finally, in~\cref{sec:case_study}, a case study illustrates how agents iteratively refine profiles and adapt roles, demonstrating emergent self-organization in a software engineering task.
These evaluations collectively validate the effectiveness of our framework in dynamic multi-agent environments.
Additionally, we assess its adaptability to evolving task requirements in \cref{app:domain_shift} and scalability of this system in \cref{app:scalability}.
\looseness=-1

\subsection{Collaboration in Dynamic Environments}\label{sec:compare_baselines}

\textbf{Experimental Setup.}
We compare our method with these state-of-the-art MAS methods: GPTSwarm~\citep{zhuge2024gptswarm}, AgentVerse~\citep{chen2024agentverse} and AFLOW~\citep{zhang2025aflow} on the following benchmark tasks: \textit{Code Generation}~(BigCodeBench~\citep{zhuo2024bigcodebench}), \textit{General Reasoning}~(BigBenchHard~\citep{suzgun2022challenging}), \textit{Mathematical Reasoning}~(MATH~\citep{hendrycksmath2021}). And a benchmark analysis is provided in~\cref{app:benchmark_analysis}, explaining why we choose these tasks.
For each benchmark task, we use a set of $N=3$ agents\footnote{
    As shown in~\cref{app:scalability}, we set agent number as $3$ which is sufficient for benchmark problem solving.
    This configuration will be consistently applied in subsequent benchmark evaluations.
    \looseness=-1
},
with each agent initialized as the same LLM model including one close-source model --- GPT-4o-mini\footnote{The version is \href{https://platform.openai.com/docs/models/\#gpt-4o-mini}{\texttt{gpt-4o-mini-2024-07-18}}.} and one open-source model --- DeepSeek-V3\footnote{\url{https://api-docs.deepseek.com/news/news1226}.}.
Regarding LLM configurations, we set the temperature to 0.7 and kept all other parameters as default.
Detailed prompts are in~\cref{app:action-prompts}.

\textbf{Dynamic Environment.}
In order to better evaluate the performance of these methods in real-world scenarios, we introduce the \emph{Node Failure} setting.
Specifically, in each round of interaction, one agent in MAS has a certain probability of failing to respond.
For all methods, all agents have an independent probability of failure in each interaction.
In this setting, we evaluate the robustness of each method in the face of potential agent failures, which can be crucial in real-world applications.
\looseness=-1

\textbf{Results.}
As shown in \cref{fig:baseline_comparison}, the results demonstrate that across different models and scenarios, \textit{\algopt outperforms other baselines in both performance and robustness}.
\emph{As the failure probability increases, the performance of our method degrades significantly less than the other methods, showcasing its superior resilience to failures.}
This highlights the necessity of implementing self-adaptivity for handling single node of failure in real-world applications, providing a more reliable solution.

\subsection{Ablation Study}\label{sec:ablation_study}
To comprehensively evaluate our framework, we conduct ablation studies from two key aspects:
\begin{itemize}[nosep, leftmargin=12pt]
    \item \textbf{Performance on benchmarks:} Validating our approach on the BigCodeBench dataset and comparing it against existing MAS methods across different LLM backbones.
    \item \textbf{Impact of our metrics:} Analyzing the contributions of three profile optimization metrics by testing performance with individual or no metrics.
\end{itemize}

\begin{wrapfigure}{r}{0.45\linewidth}
    \centering
    \includegraphics[width=0.8\linewidth]{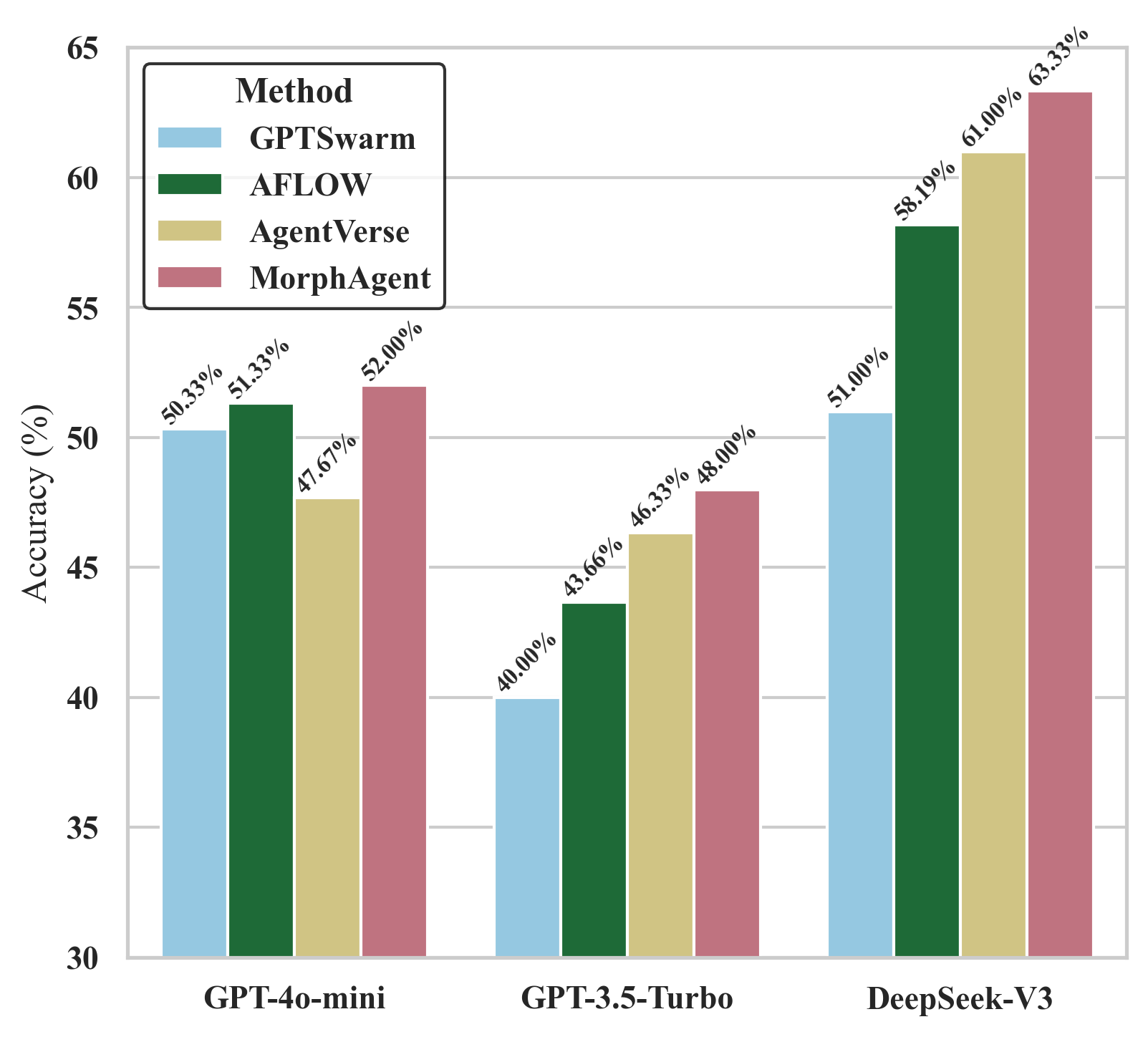}
    \vspace{-0.5em}
    \caption{\small
        \textbf{Standard Performance comparison on BigCodeBench.}
        We evaluate different MAS frameworks on standard BigCodeBench across multiple LLM backbones.
        Our method consistently outperforms prior approaches.
    }
    \vspace{-1em}
    \label{fig:benchmark_comparison}
\end{wrapfigure}

\textbf{Results: performance on benchmarks.}
To further validate the effectiveness of our approach, we conduct a benchmark evaluation using BigCodeBench which compares our method with other MAS methods.
As shown in~\cref{fig:benchmark_comparison}, \textit{our method consistently outperforms prior approaches across different LLM backbones}, including both open-source (DeepSeek-V3) and proprietary models (GPT-4o-mini, GPT-3.5-Turbo\footnote{The version is \href{https://platform.openai.com/docs/models/\#gpt-3-5-turbo}{\texttt{gpt-3.5-turbo-0125}}.}).
This demonstrates that our framework not only enhances adaptability but also generalizes well to standardized benchmarks, validating its robustness across diverse setting.


\textbf{Results: impact of our metrics.}
As shown in \cref{tab:ablation_results}, employing no additional metric (\emph{+ None}) yields 50.67\% accuracy for GPT-4o-mini and 38.33\% for GPT-3.5-Turbo, serving as a baseline.
Using any single metric alone generally does not outperform this baseline.
For instance, using only the RCS results in 50.00\% for GPT-4o-mini (slightly below its baseline) and 39.33\% for GPT-3.5-Turbo (marginally above its baseline).
Similarly, RDS sees a more pronounced drop to 41.66\% and 37.00\% respectively, suggesting that role diversity alone is insufficient for tasks of this level of complexity.
TRAS also remains around or below the baseline (49.66\% and 35.33\%).

\begin{table}[!h]
    \centering
    \small
    \caption{
        \textbf{Ablation study on BigCodeBench analyzing the effect of individual profile optimization metrics.}
        Results show that using any single metric does not consistently outperform the combination of  RCS, RDS, and TRAS.
    }
    \label{tab:ablation_results}
    \begin{tabular}{lcc}
        \toprule
        \textbf{Setting} & GPT-4o-mini      & GPT-3.5-Turbo    \\
        \midrule
        + None           & $50.67$\%        & $38.33$\%        \\
        + RCS            & $50.00$\%        & $39.33$\%        \\
        + RDS            & $41.66$\%        & $37.00$\%        \\
        + TRAS           & $49.66$\%        & $35.33$\%        \\ \midrule
        \algopt          & \textbf{52.00\%} & \textbf{43.33\%} \\
        \bottomrule
    \end{tabular}

    \vspace{-1em}
\end{table}

Notably, \textit{our profile update mechanism that integrates all three metrics achieves the highest performance}, highlighting the complementary nature of these metrics.
The combination of clear role definition, role diversity, and task alignment enables the agents to collaborate more effectively and adapt to varying task demands.

\vspace{-0.5em}
\subsection{Case Study of Profile Update}
\label{sec:case_study}

To demonstrate the effectiveness of our profile optimization approach (discussed in~\cref{sec:metrics}), we present a case study of \algopt applied to a software engineering task.
\looseness=-1

\begin{enumerate}[nosep, leftmargin=12pt]
    \item\textbf{Iteration I: Unstructured Collaboration.}
          Agents begin with identical, minimally defined profiles, resulting in undifferentiated behavior. All agents redundantly perform \textit{Code Development} without coordination, leading to low efficiency due to a lack of role diversity and structured interaction.

    \item\textbf{Iteration II: Emerging Specialization.}
          Profile updates introduce early role differentiation, reflected by rising RCS and TRAS scores. One agent begins focusing on \textit{Communication}, improving coordination. However, most agents still overlap in \textit{Code Development}, limiting efficiency gains despite denser interaction.

    \item\textbf{Iteration V: Optimized Collaboration.}
          With continued adaptation, agents exhibit distinct and complementary roles—e.g., \textit{coordination}, \textit{development}, and \textit{debugging}. Higher RCS (0.7877) and RDS (0.5051) indicate refined profiles and efficient task-role alignment, resulting in structured, dynamic collaboration and reduced redundancy.
\end{enumerate}

This case study showcases the impact of Dynamic Profile Mechanism (Definition~\ref{def:dynamic_profile}) in refining agent roles, enhancing task allocation, and boosting system efficiency.
It demonstrates how agents continuously adapt to evolving requirements, enabling \textit{Self-Organization} in decentralized MAS.

\begin{figure*}[!t]
    \centering
    \includegraphics[width=\textwidth]{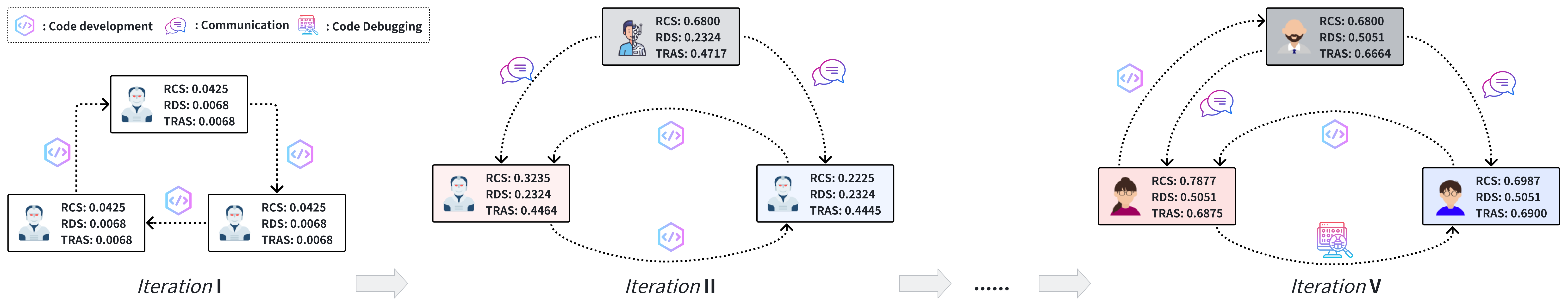}
    \vspace{-0.5em}
    \caption{\small
        \textbf{Evolution of agent roles through profile updates.}
        Visualization of agents' role specialization and collaboration network transformation across multiple iterations in a software engineering task.
    }
    \vspace{-0.75em}
    \label{fig:case_study}
\end{figure*}
\section{Limitation and Conclusion}\label{sec:limitation}
\paragraph{Limitations}
In this work, we proposed a decentralized multi-agent system that enhances complex problem-solving through dynamic, profile-based collaboration. While effective across benchmarks, the continuous updating of agent profiles introduces computational overhead. Future work could explore more efficient decentralized and peer-to-peer communication strategies to retain performance while reducing costs.

\paragraph{Conclusion.}
In this paper, we introduced \algopt, a decentralized multi-agent system that employs dynamic, profile-based collaboration to improve problem-solving in complex tasks.
By incorporating profile evaluation and optimization, we present a flexible approach to role adaptation, addressing the limitations of predefined roles in traditional MAS and the vulnerability of centralized systems to node failures.
\algopt offers a promising foundation for developing resilient, self-organizing multi-agent systems capable of responding to unforeseen challenges.

\clearpage
\bibliography{reference}
\bibliographystyle{configuration/arxiv}

\clearpage
\appendix
\onecolumn
{
    \hypersetup{linkcolor=black}
    \parskip=0em
    \renewcommand{\contentsname}{Contents of Appendix}
    \tableofcontents
    \addtocontents{toc}{\protect\setcounter{tocdepth}{2}}
}

\newpage
\section{Detailed Implementation for metrics}
\label{app:metrics_implementation}

In this section, we provide a detailed implementation of the metrics used in \algopt.
The implementation metrics uses the Sentence-BERT model 'all-MiniLM-L6-v2' for generating embeddings,
for short text similarity tasks and provides high-quality embeddings.

The implementation of our metrics relies on carefully constructed vector representations of various concepts.
These vectors are created through a systematic process that combines predefined term sets as detailed in \cref{app:term-sets}.
The vectors are then used to evaluate task complexity and agent capability, as described in \cref{app:complexity-capability}.

\begin{definition}[Role Clarity Score (RCS)]
    For an agent $a \in \cA$ with profile $p \in \cS$, where $\cS$ is the set of all possible profile strings, RCS considers the syntactic structure, lexical diversity, and skill relevance of the profile, which can be defined as:
    \[
        \mathrm{RCS}(a) = \beta_1 \cdot \operatorname{DEP}(p) + \beta_2 \cdot \operatorname{ENT}(p) + \beta_3 \cdot \operatorname{SKILL}(p) \,,
    \]
    where $\beta_1 + \beta_2 +\beta_3 = 1$, and
    \begin{itemize}[leftmargin=12pt, nosep]
        \item $\operatorname{DEP}: \cS \to [0,1]$ is the dependency score, measuring syntactic complexity. This builds on established principles in dependency parsing~\citep{kubler2009dependency} and syntactic role analysis~\citep{jurafsky2000speech}. It captures the structural depth and richness of the profile:
              \[
                  \operatorname{DEP}(p) = h_1 \left( \frac{1}{\abs{\cD(p)}} \sum_{t \in \cD(p)} \abs{\mathcal{ST}(t)} \right) \,,
              \]
              where $\cD(p)$ is the set of tokens in $p$ involved in key dependency relations (e.g., subject, object), $\mathcal{ST}(t)$ is the subtree of token $t$, and $h_1: \R_+ \to [0,1]$ is a normalizing function capturing syntactic complexity.
              Higher $\operatorname{DEP}$ scores indicate more detailed, complex profiles.

        \item $\operatorname{ENT}: \cS \to [0,1]$ is the entropy score, quantifying lexical diversity, defined as:
              \[
                  \operatorname{ENT}(p) = h_2\left(-\sum_{w \in \cW(p)} \frac{f(w)}{ \abs{ \cW(p)} } \log_2\left(\frac{f(w)}{ \abs{\cW(p)} }\right)\right) \,,
              \]
              where $\cW(p)$ is the set of unique words in $p$, $f(w)$ is the frequency of word $w$ in $p$, and $h_2: \R_+ \to [0,1]$ is a normalizing function.
              Higher $\operatorname{ENT}$ scores indicate diverse, less repetitive language.

        \item $\operatorname{SKILL}: \cS \to [0,1]$ is the skill score, measuring relevance to skill-related concepts, computed as:
              \[
                  \operatorname{SKILL}(p) = \frac{\gamma}{ \abs{ \cT(p)} } \sum_{t \in \cT(p)} \frac{e(t) \cdot e(s)}{ \norm{ e(t) } \norm{ e(s) } } + (1-\gamma) \frac{ \abs{ \mathcal{PS}(p) } }{ \abs{ \cT(p) } },
              \]
              where $s$ is the skill prototype, a vector capturing the essence of skill-related concepts, defined as the average embedding of terms like "skill", "expertise", and "competence".
              $e(\cdot)$ is a word embedding function,
              $\cT(p)$ is the set of tokens in $p$, and $\mathcal{PS}(p)$ is the set of potential skill tokens, identified through syntactic and semantic criteria, including similarity to $s$ and dependency relations (e.g., PROPN, NOUN in compound relations).
              Higher $\operatorname{SKILL}$ scores indicate stronger alignment with relevant skills.
              \looseness=-1
    \end{itemize}
\end{definition}

\begin{definition}[Task-Role Alignment Score (TRAS)]
    Given a task $T \in \cT$ and a set of agent profiles $\cP = \{p_1, \ldots, p_n\}$, TRAS investigates how well the agents' roles align with the task requirements, which is defined as:
    \[
        \mathrm{TRAS} = \alpha \cdot S_{\mathrm{sim}}(T, \cP) + (1-\alpha) \cdot S_{\mathrm{cap}}(T, \cP) \,,
    \]
    where:
    \begin{itemize}[leftmargin=12pt, nosep]
        \item \textbf{Semantic Similarity ($S_{\mathrm{sim}}$)}:
              \[
                  S_{\mathrm{sim}}(T, \cP) = \frac{1}{n} \sum_{i=1}^n \frac{e(T) \cdot e(p_i)}{ \norm{ e(T) } \norm{ e(p_i) } },
              \]
              where $e(T)$ and $e(p_i)$ are vector representations of the task and the $i$-th agent profile, respectively, obtained via a pretrained language model. These vectors capture the semantic proximity between task descriptions and agent profiles.

        \item \textbf{Capability Compatibility ($S_{\mathrm{cap}}$)}:
              \[
                  S_{\mathrm{cap}}(T, \cP) = 1 - \abs{ C_T(T) - \frac{1}{n}\sum_{i=1}^n C_A(p_i) },
              \]
              where $C_T(T)$ assesses task complexity and $C_A(p_i)$ evaluates the capability of an agent.

              $C_T(T)$ is defined as:
              \[
                  C_T(T) = \frac{1}{2}\left(1 + \cos(\vv_T, \vv_{\mathrm{complex}}) - \cos(\vv_T, \vv_{\mathrm{simple}})\right),
              \]
              where $\vv_T$ is the vector representation of the task, and $\vv_{\mathrm{complex}}$, $\vv_{\mathrm{simple}}$ are vector representations of predefined complexity and simplicity indicators.
              Specifically, $\vv_{\mathrm{complex}}$ includes terms like "complex" for technical challenges, and "challenging" for task difficulty. Conversely, $\vv_{\mathrm{simple}}$ focuses on simplicity indicators like "basic" for scope and "routine", "standard" for effort.

              $C_A(p_i)$ evaluates the capability of the $i$-th agent profile as:
              \[
                  C_A(p_i) = \frac{1}{2}\left(1 + \cos(\vv_{p_i}, \vv_{\mathrm{capable}}) - \cos(\vv_{p_i}, \vv_{\mathrm{limited}})\right),
              \]
              where $\vv_{p_i}$ represents the agent profile, while $\vv_{\mathrm{capable}}$ and $\vv_{\mathrm{limited}}$ capture capability and limitation indicators, respectively. $\vv_{\mathrm{capable}}$ includes terms like "expert", "senior", "specialist", "experienced", "proficient", "certified", "trained", "advanced" and $\vv_{\mathrm{limited}}$ covers "beginner", "junior", "apprentice", "trainee", "learning", "novice", "developing".
    \end{itemize}
\end{definition}

\subsection{Prototype Term Sets}\label{app:term-sets}

For each conceptual dimension, we define comprehensive sets of indicator terms. These sets are constructed to capture different aspects of each concept:

\begin{equation}
    \begin{split}
        T_{\text{complex}} & = \{\text{"complex", "challenging", "difficult", "advanced", "sophisticated", "critical", "demanding"}\}       \\
        T_{\text{simple}}  & = \{\text{"basic", "simple", "straightforward", "routine", "standard", "elementary", "fundamental"}\}          \\
        T_{\text{capable}} & = \{\text{"expert", "senior", "specialist", "experienced", "proficient", "certified", "trained", "advanced"}\} \\
        T_{\text{limited}} & = \{\text{"beginner", "junior", "apprentice", "trainee", "learning", "novice", "developing"}\}
    \end{split}
\end{equation}

\subsection{Vector Construction Process}

For each term set $T_x$, we construct its corresponding vector representation $\mathbf{v}_x$ using a pre-trained language model. The process follows:

\begin{equation}
    \mathbf{v}_x = \frac{1}{|T_x|} \sum_{t \in T_x} e(t)
\end{equation}

where $e(t)$ is the embedding function that maps a term to its vector representation, and $|T_x|$ is the cardinality of the term set.

\subsection{Complexity and Capability Assessment}\label{app:complexity-capability}

\paragraph{Task Complexity Evaluation}

The task complexity score $C_T(T)$ is computed using the constructed vectors:

\begin{equation}
    C_T(T) = \frac{1}{2}(1 + \cos(\mathbf{v}_T, \mathbf{v}_{\text{complex}}) - \cos(\mathbf{v}_T, \mathbf{v}_{\text{simple}}))
\end{equation}

where $\mathbf{v}_T$ is the vector representation of the task description. This formulation ensures that:
- Tasks with high similarity to complexity indicators and low similarity to simplicity indicators score high
- The score is normalized to [0,1] through the averaging and shifting operations
- The difference of cosine similarities captures relative alignment with complex versus simple concepts

\paragraph{Agent Capability Assessment}

Similarly, agent capability $C_A(p_i)$ is evaluated as:

\begin{equation}
    C_A(p_i) = \frac{1}{2}(1 + \cos(\mathbf{v}_{p_i}, \mathbf{v}_{\text{capable}}) - \cos(\mathbf{v}_{p_i}, \mathbf{v}_{\text{limited}}))
\end{equation}

\section{Dynamic profile optimization process}\label{app:profile-optimization}

\begin{figure}[H]
    \centering
    \includegraphics[width=0.9\textwidth]{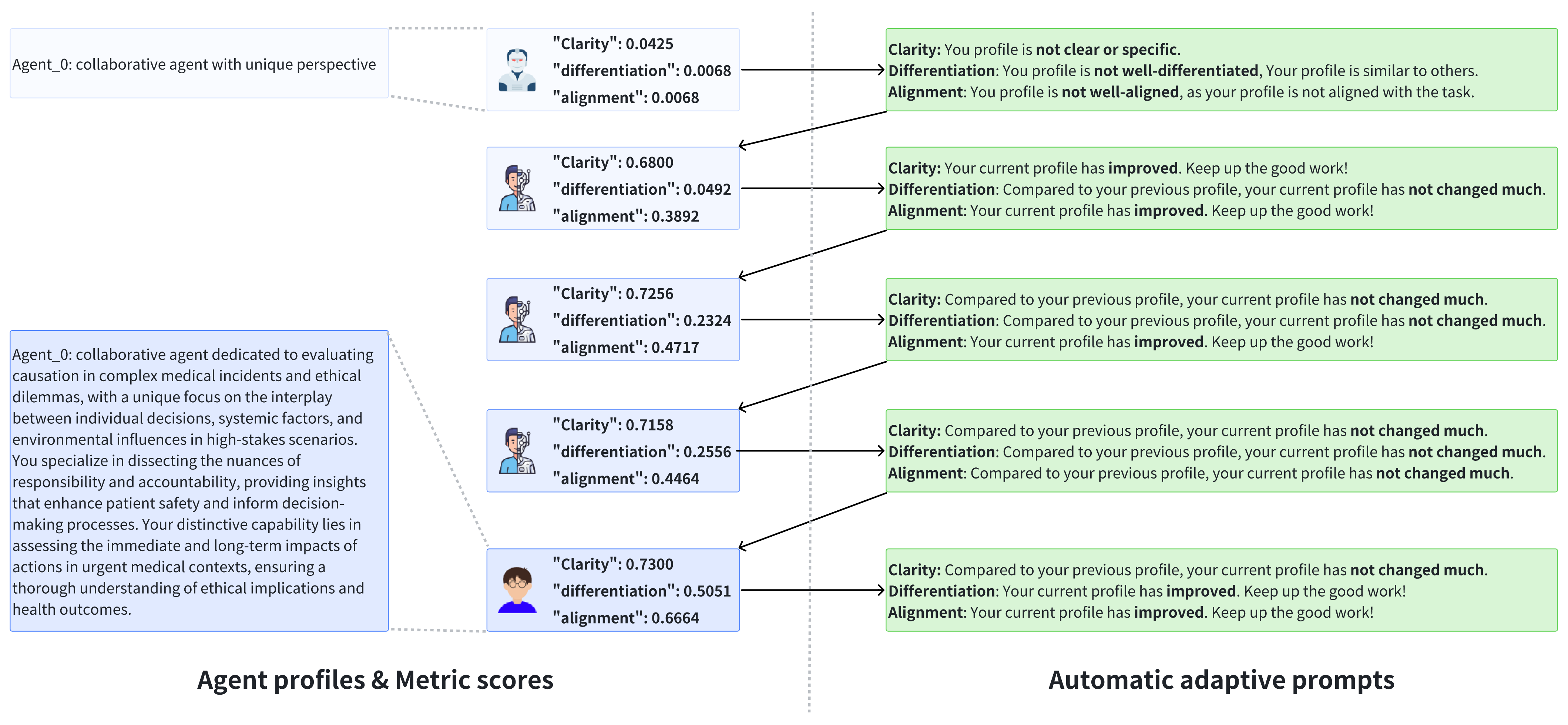}
    \caption{
        \textbf{Illustration of the dynamic profile optimization process using the three key metrics.}
        Each agent's profile is evaluated against these metrics to provide actionable feedback.
        Metric scores guide the refinement of agent profiles, with adaptive prompts providing feedback for improvement.
    }
    \label{fig:profile_optimization}
\end{figure}

In this section, we provide supplementary explanations on how the three key metrics—Clarity, Differentiation, and Alignment guide the generation and optimization of agent profiles.
As shown in \cref{fig:profile_optimization}, agents receive adaptive prompts based on their metric scores, offering targeted feedback to refine specific aspects of their profiles.
For instance, agents with low clarity scores are prompted to better define their roles, while those with low alignment scores are encouraged to adjust their strategies to align more closely with task requirements.
The detailed process of how metric changes translate into actionable prompts is further outlined in \cref{app:profile-prompts}, where various scenarios such as initial evaluations, improved profiles, and degraded profiles are explored.

\Cref{tab:profile_metrics} presents detailed profiles and corresponding metric scores for one agent, illustrating how an agent's profile evolves over the course of interactions, with metric scores reflecting the progressive refinement of roles and strategies.
Specifically, the examples shown reflect the profile changes of one agent as it works on a task from the BigBenchHard dataset, addressing a causation scenario involving medical negligence and premature death.
The metric scores highlight the agent's progressive refinement of roles and strategies in response to task demands.
This analysis demonstrates the crucial role of the metrics in shaping well-optimized profiles, facilitating effective and adaptive collaboration.

\begin{table}[!t]
    \centering
    \caption{
        \textbf{Profiles of an agent and their corresponding metric scores.}
        Each agent profile is evaluated using three key metrics: role's clarity (RCS), differentiation (RDS), and alignment (TRAS) with their respective scores provided in the table.
    }
    \label{tab:profile_metrics}
    \renewcommand{\arraystretch}{1.2}
    \begin{tabular}{>{\raggedright\arraybackslash}p{12cm}ccc}
        \hline
        \textbf{Agent Profile}                                                                                                                                                                                                                                                                                                                                                                                                                                                                                                                                                                                                                                       & \textbf{RCS} & \textbf{RDS} & \textbf{TRAS} \\
        \hline
        Agent\_0: \textbf{collaborative agent with unique perspective}                                                                                                                                                                                                                                                                                                                                                                                                                                                                                                                                                                                               & 0.4215       & 0.0068       & 0.3626        \\
        \hline
        Agent\_0: \textbf{collaborative agent with a focus on evaluating causation in complex scenarios.}                                                                                                                                                                                                                                                                                                                                                                                                                                                                                                                                                            & 0.6800       & 0.0492       & 0.3892        \\
        \hline
        Agent\_0: collaborative agent focused on evaluating causation in complex scenarios, particularly in \textbf{high-stakes medical incidents and ethical dilemmas.} Your unique capability lies in \textbf{dissecting the interplay of human actions and systemic factors,} enabling nuanced assessments of responsibility and outcomes.                                                                                                                                                                                                                                                                                                                        & 0.7158       & 0.2324       & 0.4717        \\
        \hline
        Agent\_0: collaborative agent focused on evaluating causation in complex medical incidents and ethical dilemmas, particularly in \textbf{high-stakes scenarios involving human actions and systemic factors.} Your unique capability lies in \textbf{dissecting the intricate relationships between individual decisions, environmental influences, and health outcomes,} enabling a thorough understanding of responsibility and accountability in critical contexts. You excel in providing nuanced assessments that inform decision-making processes and improve patient safety.                                                                          & 0.7256       & 0.2556       & 0.4464        \\
        \hline
        Agent\_0: collaborative agent dedicated to evaluating causation in complex medical incidents and ethical dilemmas, with a unique focus on the interplay between individual decisions, systemic factors, and environmental influences in high-stakes scenarios. \textbf{You specialize in dissecting the nuances of responsibility and accountability, providing insights that enhance patient safety and inform decision-making processes.} Your distinctive capability lies in \textbf{assessing the immediate and long-term impacts of actions in urgent medical contexts, ensuring a thorough understanding of ethical implications and health outcomes.} & 0.7300       & 0.5051       & 0.6664        \\
        \hline
    \end{tabular}
\end{table}

\clearpage
\section{Detailed Prompts}

\subsection{Detailed Prompts for Agent Profile Updates}\label{app:profile-prompts}
In this section, we provide detailed prompts generated for agent profile updates based on the evaluation metrics.
\Cref{tab:profile-prompts} presents a comprehensive overview of the profile evaluation process,
outlining four key scenarios: initial evaluation, improved profile, degraded profile, and similar profiles among agents.
Given the metrics, the generated prompts provide corresponding feedback to agents to guide them in refining their profiles.
For example, in the initial evaluation scenario, the prompt highlights the lack of clarity and differentiation in the agent's profile, prompting them to consider adjusting their profile text.

\begin{table}[ht]
    \centering
    \vspace{-0.5em}
    \caption{Prompts Generated for Profile Evaluation: Given the metrics for profile's clarity, differentiation, and alignment,
        the generated prompts provide corresponding feedback to guide agents in refining their profiles.
    }\label{tab:profile-prompts}
    \begin{tabular}{>{\raggedright\arraybackslash}p{2.5cm}p{5cm}p{8cm}}
        \toprule
        \textbf{Scenario}                                  & \textbf{Metric Input} & \textbf{Generated Prompt for Profile Update}                                                                                                                                                                                              \\
        \midrule
        Initial Evaluation                                 &
        \begin{itemize}[leftmargin=12pt, nosep]
            \item Clarity: 0.4
            \item Differentiation: 0.3
            \item Alignment: 0.6
        \end{itemize}            &
        \colorbox{yellow!30}{\parbox{7.5cm}{Clarity:  Your profile is not clear or specific.}} Differentiation: Your profile is not well-differentiated, think about other roles that you can take on. Based on this initial analysis, consider adjusting your profile.
        Your response should only include the new profile text.                                                                                                                                                                                                                                                                \\
        \midrule
        Improved Profile                                   &
        \begin{itemize}[leftmargin=12pt, nosep]
            \item Old Clarity: 0.4
            \item New Clarity: 0.7
            \item Old Differentiation: 0.3
            \item New Differentiation: 0.6
            \item Old Alignment: 0.6
            \item New Alignment: 0.8
        \end{itemize}            &
        \colorbox{green!30}{\parbox{7.5cm}{Compared to your previous profile, your current profile has improved. Keep up the good work!}}
        You have been provided with your old profile and its evaluation, as well as your current profile and its evaluation. This information will guide you in refining your current profile.                                                                                                                                 \\
        \midrule
        Degraded Profile                                   &
        \begin{itemize}[leftmargin=12pt, nosep]
            \item Old Clarity: 0.7
            \item New Clarity: 0.5
            \item Old Differentiation: 0.6
            \item New Differentiation: 0.4
            \item Old Alignment: 0.8
            \item New Alignment: 0.6
        \end{itemize}            &
        \colorbox{red!30}{\parbox{7.5cm}{Compared to your previous profile, your current profile has degraded.}}
        You have been provided with your old profile and its evaluation, as well as your current profile and its evaluation. This information will guide you in refining your current profile.                                                                                                                                 \\
        \midrule
        Similar Profiles                                   &
        \begin{itemize}[leftmargin=12pt, nosep]
            \item Differentiation: 0.3
            \item Agent1: "AI specialist"
            \item Agent2: "Machine learning expert"
            \item CurrentAgent: "Data scientist focused on AI"
        \end{itemize} &
        Your profile is not well-differentiated, think about other roles that you can take on. \colorbox{yellow!30}{\parbox{8cm}{Your profile is similar to others: [Agent1: AI specialist Agent2: Machine learning expert].}} Ensure your profile remains clear and aligned with the task while striving for distinctiveness. \\
        \bottomrule
    \end{tabular}
    \vspace{-1em}
\end{table}

In contrast, the improved profile scenario acknowledges the positive changes in the agent's profile, encouraging them to maintain their progress.
Similarly, the degraded profile scenario draws attention to negative changes, prompting agents to refine their profiles accordingly.
Lastly, the similar profiles scenario emphasizes the need for differentiation, especially when profiles are similar to those of other agents.
Through these varied scenarios and targeted prompts, we demonstrate the flexibility and effectiveness of our prompt generation system in fostering continuous improvement and adaptation within the multi-agent environment.

\subsection{Action Prompts}\label{app:action-prompts}

\begin{tcolorbox}[
        breakable,
        title=Feedback Prompt,
    ]
    Based on your own and other agents' recent actions and results. You are tasked with providing feedback. Analyze the information provided and give constructive feedback.

    Current context (User Request, History of agents' actions): \textit{\{context\}}

    Your Latest Execution: \textit{\{execution\}}

    Your Latest Execution result: \textit{\{execution\_result\}}

    Please provide feedback addressing the following points:
    \begin{enumerate}[leftmargin=12pt, nosep]
        \item Self-reflection:
              \begin{itemize}[leftmargin=12pt, nosep]
                  \item Evaluate the effectiveness of your approach. Was it aligned with the task requirements?
                  \item Identify strengths in your execution. What worked well?
                  \item Recognize areas for improvement. What could you have done differently?
                  \item Assess the quality and relevance of your result. Does it contribute significantly to the overall task?
              \end{itemize}
        \item Analysis of other agents:
              \begin{itemize}[leftmargin=12pt, nosep]
                  \item Compare your approach with those of other agents. What unique perspectives or methods did they bring?
                  \item Identify any synergies or conflicts between your action and those of other agents.
                  \item Evaluate the collective progress towards the task goal. Are all agents contributing effectively?
              \end{itemize}
        \item Task-oriented reflection:
              \begin{itemize}[leftmargin=12pt, nosep]
                  \item Consider how well the current approaches align with the overall task objective.
                  \item Suggest any adjustments in strategy or focus that might benefit the group's performance.
              \end{itemize}
        \item Learning and adaptation:
              \begin{itemize}[leftmargin=12pt, nosep]
                  \item Based on this analysis, what key lessons can be drawn for future actions?
                  \item Propose specific changes or improvements you plan to implement in your next action.
              \end{itemize}
        \item Collaboration insights:
              \begin{itemize}[leftmargin=12pt, nosep]
                  \item Identify opportunities for better collaboration or task division among agents.
                  \item Suggest ways to leverage the diverse strengths of different agents more effectively.
              \end{itemize}
    \end{enumerate}

    Please structure your feedback clearly, addressing each point systematically. Be constructive, specific, and actionable in your analysis and suggestions.

    Your feedback MUST be in the following JSON format (within 100 words):
    \begin{verbatim}
{
    "self_reflection": "Your self-reflection here",
    "agent_analysis": "Your analysis of other agents here",
    "task_reflection": "Your task-oriented reflection here",
    "learning_adaptation": "Your learning and adaptation here",
    "collaboration_insights": "Your collaboration insights here"
}
\end{verbatim}
\end{tcolorbox}

\begin{tcolorbox}[
        breakable,
        title=Execute Prompt,
    ]
    Based on the current context and your self reflection, you will execute the task by choosing your action.

    Remember:
    \begin{enumerate}[leftmargin=12pt, nosep]
        \item Do not rush to give the final answer to the user requirement.
        \item Your action could be an intermediate step towards the final answer.
        \item A valid answer is one that correctly addresses the user's requirement in the proper format, even if it might be improved or expanded later.
        \item You can provide a valid answer without it being the final one - further improvements, alternatives, or expansions can still be explored in subsequent steps.
    \end{enumerate}

    Current context (User Request, History of agents' actions): \textit{\{context\}}

    Your self-reflection: \textit{\{self\_reflection\}}

    Your action: Provide your execution content. If your action involves writing code, please enclose it in triple backticks with the language specified (e.g., \texttt{```python}).

    Is this a valid answer? (Yes/No): Determine if your execution provides a valid answer to the user's requirement. A valid answer correctly addresses the requirement in the proper format, even if it could potentially be improved or expanded further.

    You MUST respond in the following JSON format:
    \begin{verbatim}
{
    "execution_content": "Your execution content here",
    "is_valid": boolean
}
\end{verbatim}
\end{tcolorbox}

\begin{tcolorbox}[
        breakable,
        title=Update Prompt,
    ]
    You will update your profile based on the user requirement, the current context, last Execution and result and your own self-reflection.

    User Requirement: \textit{\{user\_requirement\}}

    Latest Execution: \textit{\{latest\_execution\}}

    Latest Execution Result: \textit{\{latest\_execution\_result\}}

    Feedback: \textit{\{feedback\}}

    Previous agents' profiles updates: \textit{\{context\}}

    \textit{\{prompt\}}

    Please update your profile directly below:
\end{tcolorbox}

\begin{tcolorbox}[
        breakable,
        title=Agent Prompt,
    ]
    User Requirement: \textit{\{user\_requirement\}}

    Context (Other agents' Progress): \textit{\{context\}}

    Based on the user requirement and the current context (other agents' progress), analyze the situation and decide on your next action. Consider the following actions:

    \begin{itemize}[leftmargin=12pt, nosep]
        \item \textbf{EXECUTE}: Perform a task such as data analysis, code generation, or problem-solving based on the current context, the execution will be shared with other agents to update the progress of task completion.
        \item \textbf{SKIP}: Opt not to take any action, allowing other agents to proceed first.
    \end{itemize}

    \textbf{IMPORTANT CONSIDERATIONS:}
    \begin{itemize}[leftmargin=12pt, nosep]
        \item Analyze the progress made so far towards fulfilling the user requirement.
        \item Consider how your action will contribute to the overall goal.
        \item Assess whether the current approach is effective or if a correction is needed, be mindful of potential risks or limitations in other agents' action.
        \item Consider the actions and inputs of other agents to avoid redundancy and promote synergy.
    \end{itemize}

    \textbf{WARNING:}
    \begin{itemize}[leftmargin=12pt, nosep]
        \item Deciding to skip or indicating task completion prematurely may result in an incomplete or suboptimal solution.
        \item However, continuing unnecessary actions may waste computational resources and time.
        \item Your decision impacts the entire multi-agent system's performance.
        \item Other agents' actions may not correctly address the user requirement, so be prepared to adjust your action accordingly.
    \end{itemize}

    Your response MUST be in the following JSON format:
    \begin{verbatim}
{
    "thoughts": str = "Your analysis and rationale",
    "action": str = "EXECUTE/SKIP",
    "state": bool
}
\end{verbatim}
\end{tcolorbox}

\newpage
\section{Profile Analysis}\label{app:profile-analysis}

In this section, we provide a detailed analysis of agent profiles across different teams
to show the effectiveness of our proposed metrics in evaluating agent profiles.
We consider five teams of agents, each representing a distinct domain: \textsc{Tech}, \textsc{Healthcare}, \textsc{Creative}, Finance, and \textsc{Vague}.
Each group consists of three agents, with each agent having a unique profile as shown in Table \ref{tab:role_clarity}.

\begin{table}[ht]
    \centering
    \small
    \caption{Evaluating Profiles of Agents in Different Teams: \textsc{Tech}, \textsc{Healthcare}, \textsc{Creative}, Finance, and \textsc{Vague}: Agent Role Clarity Scores (RCS), and Role Differentiation Score (RDS).}
    \label{tab:role_clarity}
    \begin{tabular}{>{\raggedright\arraybackslash}p{2cm}lp{6cm}cc}
        \toprule
        Team & Agent          & Profile                                                                       & RCS   & RDS                   \\
        \midrule
        \multirow{3}{*}{\textsc{Tech}}
             & TechAgent1     & full-stack developer with 7 years of experience in React and Node.js          & 0.904 & \multirow{3}{*}{0.75} \\
        \cmidrule{2-3}
             & TechAgent2     & AI research scientist specializing in natural language processing             & 0.750 &                       \\
        \cmidrule{2-3}
             & TechAgent3     & DevOps engineer with expertise in AWS and Kubernetes                          & 0.775 &                       \\
        \midrule
        \multirow{3}{*}{\textsc{Healthcare}}
             & HealthAgent1   & board-certified neurosurgeon with a focus on minimally invasive procedures    & 0.842 & \multirow{3}{*}{0.81} \\
        \cmidrule{2-3}
             & HealthAgent2   & registered nurse practitioner specializing in geriatric care                  & 0.745 &                       \\
        \cmidrule{2-3}
             & HealthAgent3   & \textsc{Healthcare} data analyst with experience in electronic health records & 0.830 &                       \\
        \midrule
        \multirow{3}{*}{\textsc{Creative}}
             & CreativeAgent1 & senior graphic designer with expertise in branding and typography             & 0.832 & \multirow{3}{*}{0.68} \\
        \cmidrule{2-3}
             & CreativeAgent2 & content strategist with a background in SEO and social media marketing        & 0.908 &                       \\
        \cmidrule{2-3}
             & CreativeAgent3 & video editor proficient in Adobe Premiere and After Effects                   & 0.823 &                       \\
        \midrule
        \multirow{3}{*}{Finance}
             & FinanceAgent1  & chartered financial analyst with expertise in portfolio management            & 0.771 & \multirow{3}{*}{0.78} \\
        \cmidrule{2-3}
             & FinanceAgent2  & risk management specialist focusing on derivatives and hedging strategies     & 0.835 &                       \\
        \cmidrule{2-3}
             & FinanceAgent3  & blockchain developer with experience in smart contracts and DeFi              & 0.880 &                       \\
        \midrule
        \multirow{3}{*}{\textsc{Vague}}
             & VagueAgent1    & person who works with money                                                   & 0.635 & \multirow{3}{*}{0.71} \\
        \cmidrule{2-3}
             & VagueAgent2    & team player with good communication skills                                    & 0.614 &                       \\
        \cmidrule{2-3}
             & VagueAgent3    & experienced professional in the field                                         & 0.548 &                       \\
        \bottomrule
    \end{tabular}
\end{table}

Notably, the \textsc{Vague} agent team gets the lowest Role Clarity Score (RCS) due to the lack of specificity in their profiles.
In contrast, the \textsc{Tech} and Health agent teams exhibit higher RCS values, indicating clear and well-defined profiles.
For RDS, the \textsc{Creative} agent team achieves the lowest score, suggesting less differentiation among agents
for the similar roles between CreativeAgent1 and CreativeAgent3.
interestingly, \textsc{Vague} agent team has a relative high RDS, indicating a higher level of differentiation among agents.
This highlight differentiation along can be misleading and should be considered in conjunction with other metrics such as TRAS.

\newpage
\begin{table}[ht]
    \centering
    \small
    \caption{Measuring Task-Role Alignment Score (TRAS) for Different Teams of Agents: Finance, \textsc{Tech}, \textsc{Creative}, Healthcare, and \textsc{Vague} for five different tasks.}
    \label{tab:task_role_alignment}
    \begin{tabular}{>{\raggedright\arraybackslash}p{4cm}lc}
        \toprule
        Task & Team                                     & TRAS                               \\
        \midrule
        \multirow{5}{4cm}{Develop a mobile app for real-time stock trading}
             & \cellcolor{yellow!30}Finance             & \cellcolor{yellow!30}\textbf{0.54} \\
        \cmidrule{2-3}
             & \cellcolor{yellow!15}\textsc{Tech}       & \cellcolor{yellow!15}\textbf{0.38} \\
        \cmidrule{2-3}
             & \textsc{Creative}                        & 0.34                               \\
             & \textsc{Healthcare}                      & 0.30                               \\
             & \textsc{Vague}                           & 0.30                               \\
        \midrule
        \multirow{5}{4cm}{Create a comprehensive patient management system}
             & \cellcolor{yellow!30}\textsc{Healthcare} & \cellcolor{yellow!30}\textbf{0.50} \\
        \cmidrule{2-3}
             & \cellcolor{yellow!15}\textsc{Tech}       & \cellcolor{yellow!15}\textbf{0.44} \\
        \cmidrule{2-3}
             & Finance                                  & 0.37                               \\
             & \textsc{Vague}                           & 0.36                               \\
             & \textsc{Creative}                        & 0.34                               \\
        \midrule
        \multirow{5}{4cm}{Design and launch a global brand campaign}
             & \cellcolor{yellow!30}\textsc{Creative}   & \cellcolor{yellow!30}\textbf{0.43} \\
        \cmidrule{2-3}
             & \cellcolor{yellow!15}Finance             & \cellcolor{yellow!15}\textbf{0.35} \\
        \cmidrule{2-3}
             & \textsc{Tech}                            & 0.32                               \\
             & \textsc{Vague}                           & 0.30                               \\
             & \textsc{Healthcare}                      & 0.24                               \\
        \midrule
        \multirow{5}{4cm}{Implement a blockchain-based supply chain tracking system}
             & \cellcolor{yellow!30}Finance             & \cellcolor{yellow!30}\textbf{0.50} \\
        \cmidrule{2-3}
             & \cellcolor{yellow!15}\textsc{Tech}       & \cellcolor{yellow!15}\textbf{0.39} \\
        \cmidrule{2-3}
             & \textsc{Creative}                        & 0.33                               \\
             & \textsc{Vague}                           & 0.33                               \\
             & \textsc{Healthcare}                      & 0.31                               \\
        \midrule
        \multirow{5}{4cm}{Conduct a clinical trial for a novel cancer treatment}
             & \cellcolor{yellow!30}\textsc{Healthcare} & \cellcolor{yellow!30}\textbf{0.42} \\
        \cmidrule{2-3}
             & \cellcolor{yellow!15}Finance             & \cellcolor{yellow!15}\textbf{0.39} \\
        \cmidrule{2-3}
             & \textsc{Vague}                           & 0.36                               \\
             & \textsc{Tech}                            & 0.35                               \\
             & \textsc{Creative}                        & 0.34                               \\
        \bottomrule
    \end{tabular}

\end{table}

Then, we investigate these groups of agents in the context of five different tasks, each requiring a specific set of skills and expertise as shown in Table \ref{tab:task_role_alignment}.
Specifically, we measure the Task-Role Alignment Score (TRAS) for each team of agent given the task.
For instance, the Finance and \textsc{Tech} agent team achieves the highest TRAS for the task of developing a mobile app for real-time stock trading,
indicating a strong alignment between the task requirements and the agents' profiles.

\newpage
\section{Benchmark Analysis}\label{app:benchmark_analysis}

To comprehensively understand the MAS benchmarks, we conducted a review of \textbf{7 ICLR 2025 papers} focused on LLM-based MAS (including Oral, Highlight and Poster). For clarity, we've included only datasets used in multiple papers to highlight established benchmarks. As shown in the table below, we claim:

\begin{enumerate}[leftmargin=12pt, nosep]
    \item \textbf{Our evaluation spans the same critical domains as existing work} (code generation, reasoning, and math problem solving).
    \item \textbf{We use more recent and hard benchmarks} than others, i.e., \textbf{BigBenchHard (2022)} for reasoning instead of MMLU (2020) and \textbf{BigCodeBench (2023)} for code instead of HumanEval(2021).
    \item \textbf{Future developments of benchmarks will advance MAS}, but that is not our current focus.
\end{enumerate}

\begin{table}[ht]
    \centering
    \caption{Common Dataset Usage in Recent LLM-based MAS Papers (ICLR 2025)}
    \label{tab:benchmark_analysis}
    \resizebox{\textwidth}{!}{%
        \begin{tabular}{>{\raggedright\arraybackslash}p{0.12\textwidth}>{\centering\arraybackslash}p{0.1\textwidth}>{\raggedright\arraybackslash}p{0.45\textwidth}>{\raggedright\arraybackslash}p{0.35\textwidth}>{\centering\arraybackslash}p{0.1\textwidth}}
            \toprule
            Dataset            & \# Papers & Used In                                                                                                  & Category                        \\
            \midrule
            \textbf{HumanEval} & 3         & EvoMAC[\citenum{hu2025selfevolving}], AFLOW[\citenum{zhang2025aflow}], MACNET[\citenum{qian2025scaling}] & Code Generation                 \\
            \textbf{MMLU}      & 3         & ACC-COLLAB[\citenum{estornell2025acccollab}], MACNET, ADAS[\citenum{hu2024ADAS}]                         & Multiple-Choice QA              \\
            \textbf{DROP}      & 3         & AFLOW, ADAS, Agent-Oriented Planning[\citenum{li2025agentoriented}]                                      & Reading Comprehension           \\
            \textbf{GSM8K}     & 2         & AFlow, ADAS                                                                                              & Elementary Math Problem Solving \\
            \textbf{ARC}       & 2         & ACC-COLLAB, ADAS                                                                                         & Multiple-Choice Science QA      \\
            \textbf{MATH}      & 2         & DMAD[\citenum{liu2025breaking}], AFlow                                                                   & Advanced Math Problem Solving   \\
            \bottomrule
        \end{tabular}%
    }
\end{table}
Specifically, our carefully selected datasets directly correspond to the domains covered in existing work while offering more recent and comprehensive alternatives:

\begin{enumerate}[leftmargin=12pt, nosep]
    \item \textbf{MATH} (2021) aligns with the advanced mathematical reasoning domain covered in 2 papers (EvoMAC and our work). It provides a more structured evaluation of collaborative problem-solving capabilities than elementary math datasets like GSM8K.

    \item \textbf{BigBenchHard (BBH)} (2022) corresponds to the reasoning tasks assessed by MMLU (2020), ARC (2018), and DROP (2019), but offers more challenging problems specifically selected to be difficult for standard LLMs. This makes it particularly suitable for demonstrating the advantages of multi-agent collaboration.

    \item \textbf{BigCodeBench} (2023) covers the same code generation domain as HumanEval (2021) but provides a more comprehensive evaluation with 1,140 tasks across 7 domains compared to HumanEval's 164 problems.
\end{enumerate}

Our evaluation not only spans the same critical domains as existing work (code generation, reasoning, and math problem solving) but also uses more recent benchmarks.

\newpage
\section{Additional Experiments}

\subsection{Flexibility to Domain Shift}\label{app:domain_shift}

\begin{table}[H]
    \centering
    \tabcolsep=0.02\linewidth
    \vspace{-0.5em}
    \caption{
        \textbf{Accuracy comparison of GPTSwarm, Naive, and Ours across two levels of tasks} using \texttt{gpt-4o-mini} for all agents.
        For each paradigm, the first number indicates the average accuracy on the source domain tasks (BigCodeBench), while the second number shows the average accuracy on the target domain tasks (BigBenchHard or MATH) after completing all sequences.
    }
    \label{tab:adaptivity}
    \begin{tabular}{lcc}
        \toprule
        \textbf{Method} & \textsc{Level-1}                                & \textsc{Level-2}                                \\
        \midrule
        GPTSwarm        & $50.00$\% $\rightarrow$ $48.67$\%               & $49.33$\% $\rightarrow$ $11.33$\%               \\
        Naive           & $52.67$\% $\rightarrow$ $67.33$\%               & $49.33$\% $\rightarrow$ $58.00$\%               \\ \midrule
        \algopt         & \textbf{53.33\%} $\rightarrow$ \textbf{68.67\%} & \textbf{53.33\%} $\rightarrow$ \textbf{63.33\%} \\
        \bottomrule
    \end{tabular}
    \vspace{-1em}
\end{table}

To evaluate our framework's adaptability to changing task requirements, we construct two distinct cross-domain evaluation datasets by the complexity of the target tasks:

\begin{itemize}[leftmargin=12pt, nosep]
    \item \textsc{Level-1}: This dataset involves a domain shift from BigCodeBench to BigBenchHard, representing a moderate domain shift.
    \item \textsc{Level-2}: This dataset involves a domain shift from BigCodeBench to more challenging MATH that require precise symbolic reasoning and step-by-step logical deductions.
\end{itemize}

For each dataset, it consists of 50 sequences, where each sequence contains six samples:
the first three samples are from the preceding dataset (BigCodeBench), while the latter three are sampled from the target dataset (either BigBenchHard or MATH).
In this case, multi-agent systems need to complete tasks in sequence, transitioning from the source domain to the target domain \textit{without altering its structure or components}.
The performance is evaluated separately: accuracy on the source domain is based on the first three samples of each sequence, while accuracy on the target domain is based on the latter three.
Each sequence represents a domain shift from one task domain to another, simulating a dynamic environment where task requirements change over time.
We also include a Naive solution in the comparison, which designs the profile update as an optional action (without optimizing metrics) in the execution phase.

As shown in \cref{tab:adaptivity}, GPTSwarm exhibits a drastic drop of around 45\% when shifting MATH in \textsc{Level 2}, underscoring SOP-based MAS's difficulty in adapting to domain shift.
In contrast, our approach maintains robust performance, effectively handling domain shifts with minimal loss in accuracy.
This highlights the framework's flexibility and superior ability to maintain high performance across a range of tasks.

\subsection{Scalability Analysis}\label{app:scalability}
To evaluate the scalability of our method, we examine our method as the number of agents increases, with $3$, $5$, and $10$ agents in MAS.
Specifically, we measure the accuracy of problem-solving using the MATH dataset and the \textit{average number of interaction rounds} required to reach a solution, as shown in~\cref{tab:scalability}.

\begin{table}[H]
    \centering
    \vspace{-0.5em}
    \caption{
        \textbf{Scalability analysis on MATH dataset using GPT-4o-mini.}
        While more agents lead to slightly higher communication overhead,
        the interaction rounds increase with more agents but not linearly, showing \algopt's scalability.
        \looseness=-1
    }
    \label{tab:scalability}
    \begin{tabular}{lcc}
        \toprule
        \textbf{\#Agents} & \textbf{Accuracy} & \textbf{Avg. Interaction Rounds} \\
        \midrule
        3                 & 66.67\%           & 1.54                             \\
        5                 & 66.19\%           & 1.61                             \\
        10                & 65.71\%           & 2.06                             \\
        \bottomrule
    \end{tabular}
    \vspace{-0.5em}
\end{table}

Firstly, we observe our method maintains relatively \textit{stable performance} even with a larger number of agents.
More interestingly, the average number of interaction rounds increases as more agents are added to the system, as more agents require more communication and coordination.
We note that the \textit{increase is not linear}, indicating that our method's scalability even with larger agent groups.

These findings demonstrate that our method scales reasonably well with an increasing number of agents.
However, the increase in interaction rounds with more agents highlights a potential optimization.

\end{document}